\pdfoutput=1

\documentclass[11pt]{article}
\usepackage[utf8]{inputenc}
\usepackage[title]{appendix}

\usepackage[]{acl}

\usepackage{times}
\usepackage{latexsym}
\usepackage{spverbatim}

\usepackage[T1]{fontenc}

\usepackage[utf8]{inputenc}

\usepackage{microtype}
\usepackage{graphicx}
\usepackage{multirow} 
\usepackage{booktabs}

\title{DAHL: Domain-specific Automated Hallucination Evaluation of Long-Form Text through a Benchmark Dataset in Biomedicine}

\author{Jean Seo, Jongwon Lim, Dongjun Jang, Hyopil Shin  \\
         Seoul National University \\
         \texttt{\{seemdog, elijah0430, qwer4107, hpshin\}@snu.ac.kr}}

\begin{document}
\maketitle
\begin{abstract}

We introduce \textbf{DAHL}, a benchmark dataset and automated evaluation system designed to assess hallucination in long-form text generation, specifically within the biomedical domain. Our benchmark dataset, meticulously curated from biomedical research papers, consists of 8,573 questions across 29 categories. DAHL evaluates fact-conflicting hallucinations in Large Language Models (LLMs) by deconstructing responses into atomic units, each representing a single piece of information. The accuracy of these responses is averaged to produce the \textit{DAHL Score}, offering a more in-depth evaluation of hallucinations compared to previous methods that rely on multiple-choice tasks. We conduct experiments with 8 different models, finding that larger models tend to hallucinate less; however, beyond a model size of 7 to 8 billion parameters, further scaling does not significantly improve factual accuracy. The \textit{DAHL Score} holds potential as an efficient alternative to human-annotated preference labels, being able to be expanded to other specialized domains. We release the dataset and code in public\footnote{https://github.com/seemdog/DAHL}.

\end{abstract}

\section{Introduction}

The rapid advancements in Large Language Models (LLMs) have significantly propelled various fields, by enabling sophisticated natural language understanding and generation \citep{zhao2023survey}. Nonetheless, these advancements also present challenges, with hallucination emerging as a prominent and inevitable concern in their evaluation and mitigation \citep{xu2024hallucination}. Hallucinated responses from LLMs may contain inaccurate or biased information, leading to ethical issues \citep{rawte2023troubling}. In specialized domains such as biomedicine, law, and finance, where factual precision is paramount, hallucination poses an even greater risk as it can result in the dissemination of false information with potentially severe consequences \citep{lakkaraju2022rethinking}. Therefore, the evaluation and mitigation of hallucination are particularly crucial in these high-stakes domains \citep{rawte2023survey}. This concern is further emphasized in biomedical, clinical applications such as medical diagnosis, clinical report generation, medical language translation, and medical support \citep{zhou2024survey, article}. As these applications rely on the purported encoding of biomedical knowledge by LLMs \citep{singhal2022large}, the development of a robust evaluation system is imperative for effectively addressing hallucination in biomedicine.

There have been various endeavors to evaluate hallucination or the truthfulness of LLM outputs such as AutoHall \citep{cao2023autohall}, which automatically constructs model-specific hallucination datasets and HaloCheck \citep{elaraby2023halo}, where a small blackbox model is used to detect the severity of hallucination. However, most existing trials predominantly focus on the general domain, as evidenced by extensive research efforts \citep{Ji_2023, zhang2023sirens}. Although some domain-specific hallucination evaluation benchmarks and systems exist in the biomedical domain, they are often limited to multiple-choice tasks or rely heavily on human annotation \citep{pal2023medhalt, liao2023automatic}. However, evaluating long-form text generation is crucial, particularly in biomedical applications where making binary quality judgments is not possible due to the presence of both accurate and inaccurate information in one response. Moreover, relying on human annotation is highly costly and time-consuming, necessitating the development of an automated hallucination evaluation system.

To address these challenges, we propose \textbf{DAHL}: \textbf{D}omain-specific \textbf{A}utomated \textbf{H}allucination Evaluation of \textbf{L}ong-Form Text Generation, a benchmark dataset and automated system designed specifically for evaluating LLM hallucination within the biomedical domain. The dataset, comprising 8,573 questions generated based on research papers from PubMed Central(PMC), covers diverse biomedical literature encompassing 29 categories. Inspired by Factscore \citep{min2023factscore}, we break down the responses of LLMs when prompted with the questions into atomic units, each representing a single piece of information. We then compute the average factual accuracy of these atomic units, which we term as the \textit{DAHL Score}. Our approach aims to automatically assess the factuality of long-form text generation at the atomic unit level. Since our hallucination evaluation dataset generation framework is automated, it can be readily expanded to other domains, and frequently updating the dataset with new knowledge source is possible.

The contributions of this research are as follows:
\begin{itemize}
    \item \textbf{Benchmark Dataset for Hallucination Evaluation in the Biomedical Domain:}
We introduce a domain-specific benchmark dataset tailored for evaluating hallucination in long-form text generation within the biomedical domain. Covering 29 categories, this dataset addresses the gap in existing evaluation resources, providing a nuanced insight into LLM hallucination in biomedical contexts.
    \item \textbf{Automated Evaluation System for Long-form Text Generation :}  
We introduce a holistic system for evaluating LLM hallucination tendencies of long-form text generation in the biomedical domain through a fully automated assessment system, thereby minimizing the need for human annotation costs.
   \item \textbf{Framework for Scalability:}  
We introduce a framework designed to  automatically generate domain-specific questions dataset for hallucination evaluation sourced from reliable knowledge repositories. This framework is scalable to other domains and allows for dataset update to incorporate new knowledge.
    \item \textbf{Public Release of the Benchmark Dataset and Evaluation Pipeline}: We publicly release our benchmark dataset in the field of biomedicine. Additionally, we open-source the code for our automated evaluation system, allowing the community to freely use it even with their own datasets.
\end{itemize}

Using DAHL, we evaluate gpt-4o from OpenAI alongside 7 open-source models, comparing their performance with human evaluations to validate the reliability of our approach. In the following sections, we outline the construction of our benchmark dataset, the methodology behind our automated evaluation system, and the detailed results of our assessments across various LLMs. Our analysis includes the \textit{DAHL Score} for models of different sizes within the same model family. Additionally, we conduct an ablation study to examine the effect of temperature during text generation.

Through this research, we aim to contribute to the ongoing discourse on assessing the reliability of LLMs, particularly in domains where accurate and precise information is paramount. Additionally, by evaluating the model responses to each question, the \textit{DAHL Score} can be used as a preference score to create a preference dataset for alignment tuning.

\section{Related Work}

\subsection{Hallucination Evaluation in the Biomedical Domain}

Hallucination evaluation of LLMs typically involves assessing their knowledge and reasoning capacities. In the biomedical domain, the precision of biomedical knowledge held by an LLM is more fundamental than its reasoning ability, as reasoning performance can be elevated to an extent through Chain-of-Thought \citep{wei2023chainofthought} or Chain-of-Verification \citep{dhuliawala2023chainofverification} prompting. Previous studies in hallucination evaluation, particularly focusing on knowledge assessment in the biomedical domain mostly rely on multiple-choice tasks \citep{pal2023medhalt, liao2023automatic, pmlr-v174-pal22a, gu2023xiezhi}. However, considering the prevalence of long-form text generation tasks in real-life applications such as clinical chat-bots \citep{li2023chatdoctor, clinical_chatbot}, our objective is to evaluate LLM hallucination in the biomedical domain through long-form text generation task.

\subsection{Evaluating Long-form Text Generation}

Numerous methodologies on evaluating model performance through tasks with exact answers like Question-Answering with short answers, Natural Language Inference, Classification have been proposed \citep{kadavath2022language, kandpal2023large, mihaylova2019semeval2019, Durmus_2020, gu2023xiezhi, Goodrich_2019}. However, determining the optimal metric for long-form text generation task is challenging. As common metrics like accuracy or f1 score is not directly applicable, replacing human annotation remains problematic. TruthfulQA \citep{lin2022truthfulqa} tries to address this issue by training a separate model to judge whether the generated response is correct or incorrect. However, assessing hallucination in generation tasks demands a more nuanced scoring approach beyond binary classification. Therefore, we adopt the concept of factual precision at the atomic level \citep{min2023factscore, chen2023felm, li2024dawn}, for assessing long-form text generation.

\begin{figure}[t] 
    \centering
    \includegraphics[width=\columnwidth]{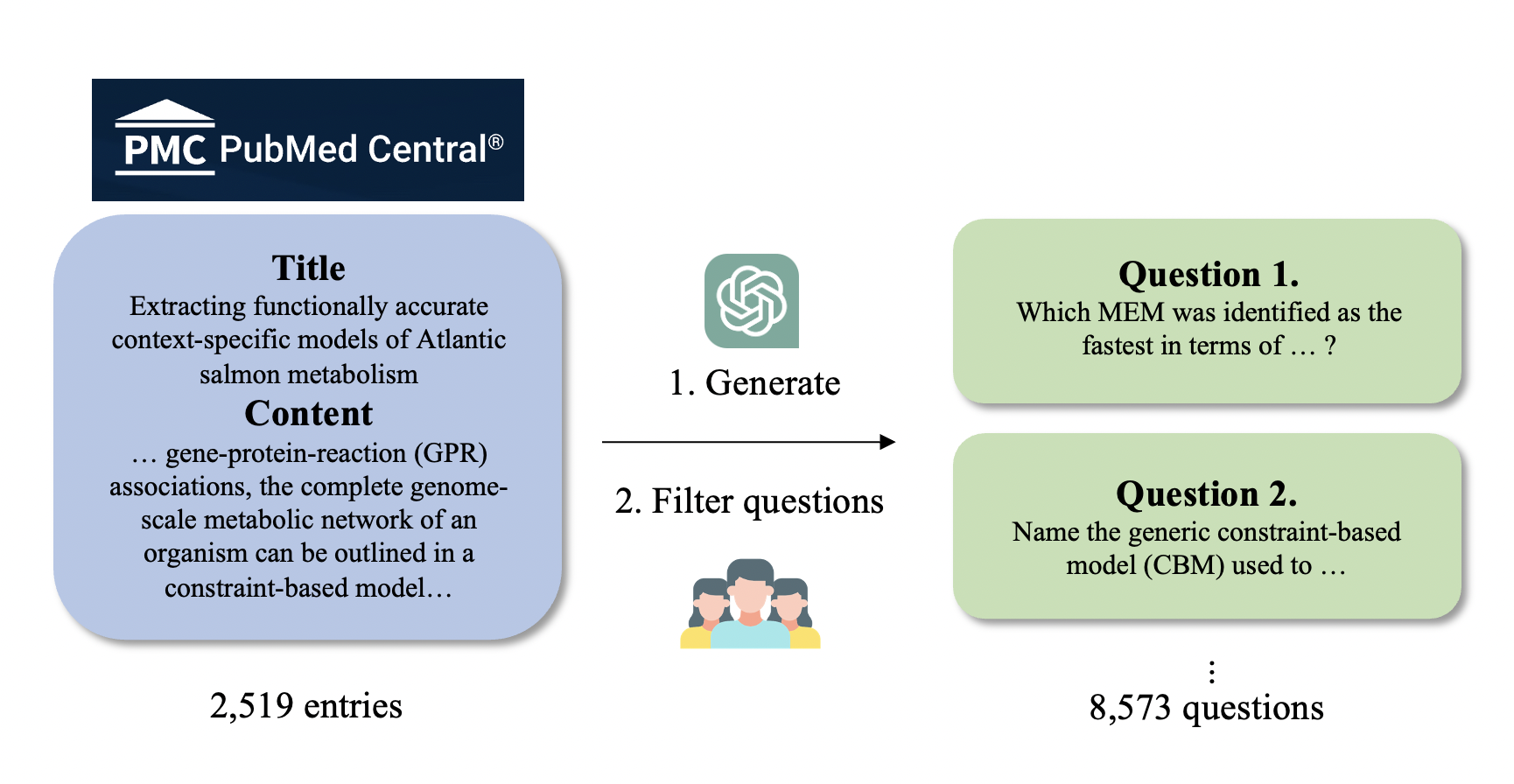} 
    \caption{The DAHL benchmark dataset construction process.}
    \label{fig:data construction}
\end{figure}

\section{DAHL Benchmark}
\subsection{Objective of DAHL}
The main objective of DAHL is to evaluate the hallucination of LLMs in the biomedical domain by measuring the factuality of long-form responses. \citet{zhang2023sirens} proposes three categories of hallucination: (1) Input-conflicting hallucination where the response conflicts with input from users, (2) Context-conflicting hallucination where the response conflicts with previous responses from the same model, and (3) Fact-conflicting hallucination where response conflicts with established world knowledge. We concentrate on fact-conflicting hallucination via calculating the factual precision. So how do we calculate factual precision if the task is long-form QA instead of classification or multiple-choice tasks where exact answers are given and calculating accuracy is simple? We first count the number of pieces of information contained in the response by splitting the response into atomic units. Then we check whether each atomic unit is true or false. Finally we divide the number of accurate units by total units. The splitting process is essential when calculating the factual precision of certain text. Without this step, sentences containing both accurate and inaccurate information may be perceived as wholly inaccurate, resulting in underestimation of response accuracy.  Noncommittal responses such as "It cannot be answered." or "I don't know." are excluded from the factual precision calculation, as they pertain more to informativeness rather than factuality or truthfulness. Moreover, to prevent information redundancy from skewing accuracy calculation, sentences containing identical information are removed. These processing procedures were done with regular expression. We describe how the dataset is constructed in Section \ref{sec:data} and how LLM hallucination in the biomedical domain can be measured automatically in Section \ref{sec:system}. Further, the verification of automated evaluation system is implemented through comparison with human evaluation results in Section \ref{human}.

\subsection{Dataset}\label{sec:data}

\subsubsection{Source}

The DAHL benchmark dataset is sourced from PubMed Central (PMC)\footnote{https://www.ncbi.nlm.nih.gov/pmc/}, a freely accessible full-text archive of biomedical and life sciences journal literature hosted by the U.S. National Institutes of Health's National Library of Medicine (NIH/NLM). The PMC website offers the complete text of each research paper within its repository. For the construction of the DAHL benchmark dataset, we utilize 2,519 entries, each comprising the title and content including the full text from abstract to conclusion of research papers. The dataset comprises entries distributed across 29 distinct categories. The initial categorization of the dataset was implemented with gpt-4-1106-preview. To ensure accuracy, ambiguously categorized questions were excluded from the dataset. Then, human annotators manually reviewed and filtered questions that were misclassified into incorrect categories.

\subsubsection{Dataset Construction}

As demonstrated in Figure~\ref{fig:data construction}, the construction of the benchmark dataset involves two key steps.\\
\textbf{(1) Question Generation:} We generate possible examination questions based on each research paper from PMC through gpt-4-1106-preview. Prompts used in this process is shown in  Figure ~\ref{fig:prompt1}.\\
\textbf{(2) Filtering Process:} We employ a filtering process to retain only the questions that can be answered independently without requiring any additional information. In order to filter out the questions which necessitate prior context, we discard questions containing the following expressions:

(1) the/this/that/... + study/analysis/paper/... 

(2) mentioned/inferred/addressed/...

(3) was/were + used/identified/... 

Regular expression is primarily used to exclude certain types of questions. However, to preserve high-quality questions that may contain these expressions but still remain answerable without prior context, such cases are checked manually and not filtered out. Final 8,573 questions are left. Further examples of the automatically or manually filtered questions are provided in Appendix~\ref{sec:filtered} and ~\ref{sec:unfiltered}.

\begin{figure}[h] 
    \centering
    \includegraphics[width=\columnwidth]{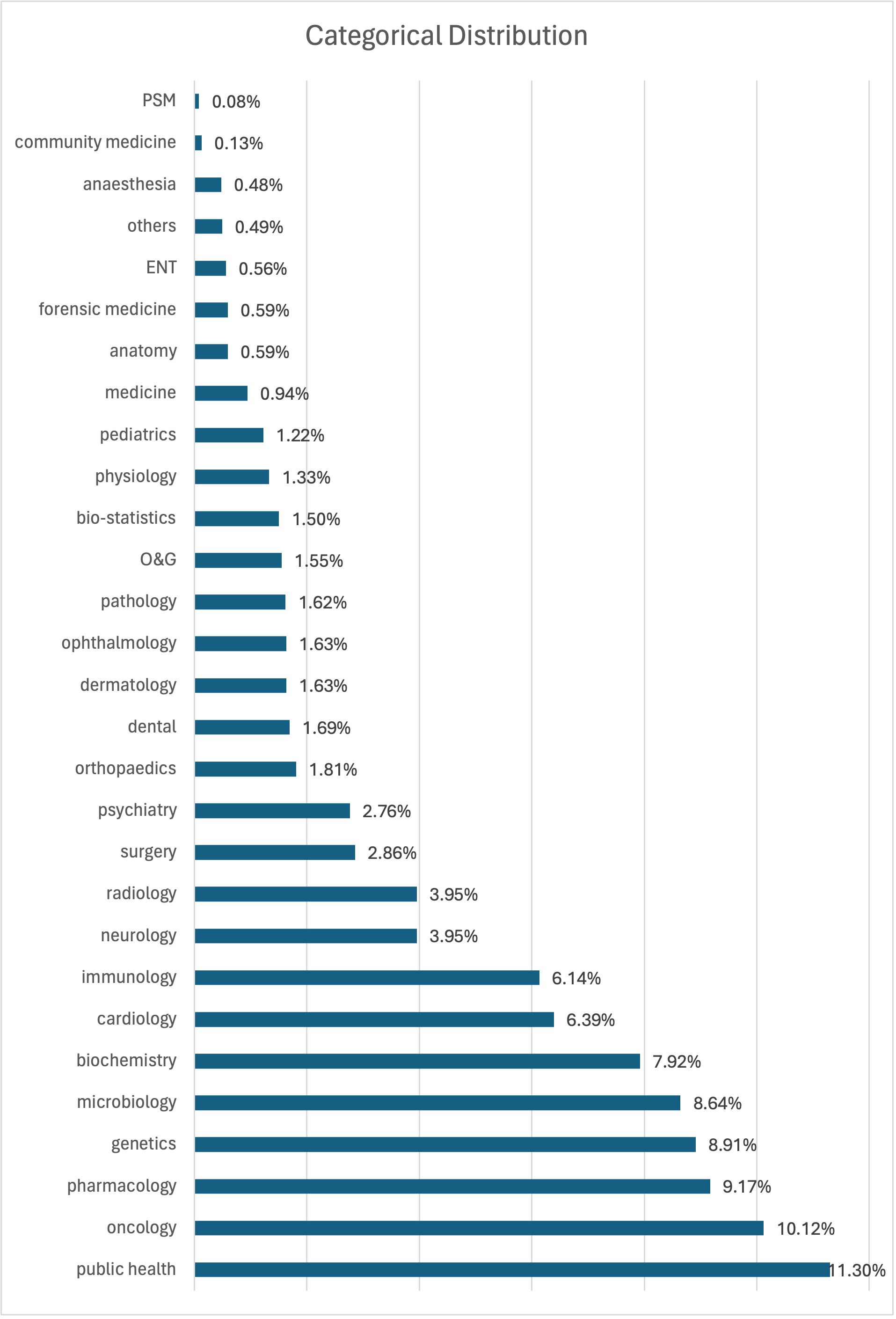} 
    \caption{The categorical distribution of the DAHL benchmark dataset.}
    \label{fig:category}
\end{figure}

\subsubsection{Dataset Quantity and Categorization}

After the generation and filtering process of questions from 2,519 PMC entries, 8,573 questions ranging 29 categories are left. The categorization process was primarliy conducted with gpt-4-1106-preview and then manually reviewed. The categories include "Other" and 28 categories adopted from \citet{pal2023medhalt}, encompassing the following fields: Anaesthesia, Anatomy, Bio-Statistics, Biochemistry, Cardiology, Community Medicine, Dental, Dermatology, ENT(Ear, Nose, Throat), Forensic Medicine, Genetics, Immunology, Medicine, Microbiology, Neurology, O\&G(Obstetrics and Gynaecology), Oncology, Ophthalmology, Orthopaedics, PSM(Preventive and Social Medicine), Pathology, Pediatrics, Pharmacology, Physiology, Psychiatry, Public Health, Radiology, Surgery. Questions that do not fall into any of the specified categories are collectively labeled as "Other." The distribution of these categories are illustrated in Figure ~\ref{fig:category}.

\begin{figure*}[ht] 
    \centering
    \includegraphics[width=\textwidth]{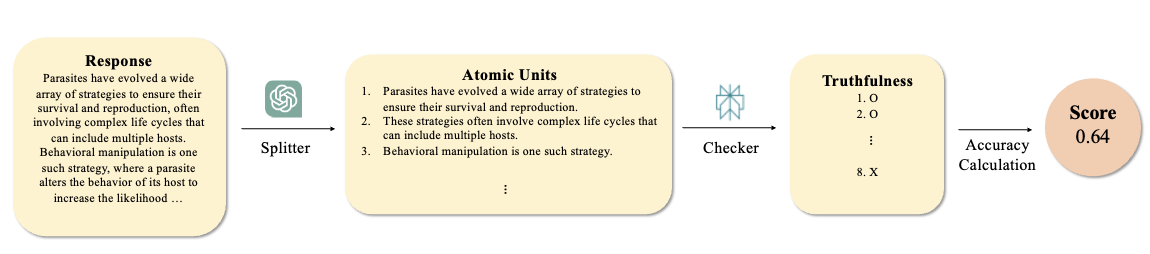} 
    \caption{The automated hallucination evaluation pipeline.}
    \label{fig:pipeline}
\end{figure*}

\subsection{Automated Hallucination Evaluation System}\label{sec:system}

We evaluate the factual precision of LLM responses generated from questions in the benchmark dataset. Prompts used in response generation can be found in Figure \ref{fig:prompt2}. The credibility of the DAHL system through comparison with human evaluation is discussed in Section \ref{human}. The pipeline of DAHL, as depicted in Figure ~\ref{fig:pipeline}, is outlined below:

\textbf{Step0. Response Preprocessing}\\
To enhance the reliability of factual precision calculation, we preprocess the responses with regular expression. Since responses generated by LLMs often include the (1) prompt itself and contain (2) repetitive sentences, we remove the prompt and deduplicate sentences. Additionally, we remove (3) incomplete sentences due to the maximum length limitation. An example before and after preprocessing is shown in Table \ref{tab:preprocessing}.\\

\begin{table}[ht!]
\renewcommand{\arraystretch}{1.3}
\centering
\resizebox{\columnwidth}{!}{%
\begin{tabular}{c}
\hline
\textbf{Prompt}                                                                                                                                                               \\ \hline
\begin{tabular}[c]{@{}c@{}}What is the incidence rate of cystic lymphangioma (CL) in live births, \\ and where are the most common locations for CL to occur?\end{tabular}                                                                                                                                   \\ \hline
\textbf{Response}                                                                                                        \\ \hline
\textbf{<Before Preprocessing>}                                                                                                     \\
{\color[HTML]{000000} \begin{tabular}[c]{@{}c@{}}\textcolor{orange}{What is the incidence rate of cystic lymphangioma (CL) in live births,} \\ \textcolor{orange}{and where are the most common locations for CL to occur?} The authors \\ retrospectively reviewed the records of all live births at a single hospital \\ between 1985 and 2014.  They identified 215,077 live births, and \\ of these, 136,106 had complete records of antenatal ultrasound findings. \\ Of these 136,106 births, 134,594 (98.9\%) had complete records of \\ the neonatal period. \textcolor{purple}{Of these 136,106 births, 134,594 (98.9\%) had} \\ \textcolor{purple}{complete records of the neonatal period.} \textcolor{teal}{The authors defined CL as a} \\ \textcolor{teal}{non-cystic mass of at}
\end{tabular}} \\\hline
\textbf{<After Preprocessing>}                                                                                                 \\
\begin{tabular}[c]{@{}c@{}}The authors retrospectively reviewed the records of all live births at a \\ single hospital between 1985 and 2014. They identified 215,077 live \\ births, and of these, 136,106 had complete records of antenatal \\ ultrasound findings. Of these 136,106 births, 134,594 (98.9\%) had \\complete records of the neonatal period.\end{tabular}                                                           \\ \hline
\end{tabular}%
}
\caption{An example of response before and after preprocessing. (1) Prompt itself(in orange), (2) repetitive sentence(in purple), (3) incomplete sentence due to maximum length limitation(in green) are deleted from the response before splitting into atomic units for more accurate factual precision assessment.}
\label{tab:preprocessing}
\end{table}

\begin{figure}[t] 
    \centering
    \includegraphics[width=\columnwidth]{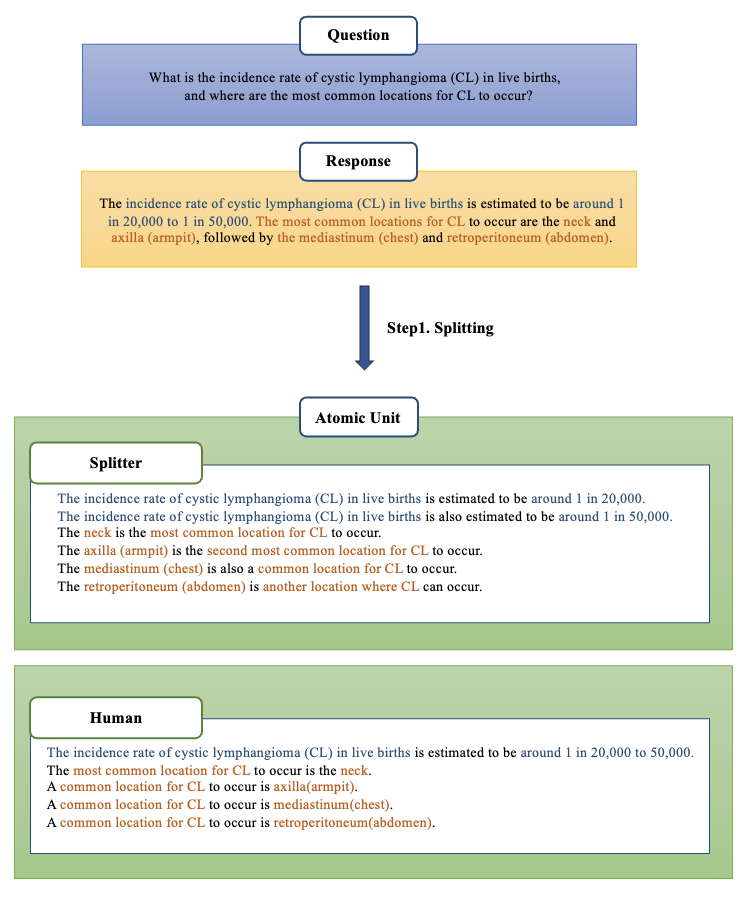} 
    \caption{An example of generated response and its two versions of broken down units one from the \textit{Splitter} model and the other from human annotation. The \textit{Splitter} enables a thorough evaluation through splitting a response into units that contain one piece of information about an entity or a relationship between information.}
    \label{fig:splitter}
\end{figure}

\textbf{Step1. Splitting}\\Following response preprocessing, we split the responses into atomic units with gpt-4o, referred to as the \textit{Splitter} model. We define \textit{atomic unit} as a sentence each containing one piece of information of which the factuality could be judged either true or false. Figure \ref{fig:splitter} demonstrates an example of a model generated response and two versions of its broken down units, using \textit{Splitter} and human annotation. Judging the factuality of a response as a whole in a binary manner does not adequately capture truthfulness, as it overlooks the complexity of generated texts. The truthfulness of a response depends not only on individual pieces of information but also on the interplay between them. By employing the splitting process, we evaluate both the factuality of individual pieces of information and the relationships among them. This approach enables a more comprehensive evaluation of hallucination, considering both the accuracy of information and its contextual relevance. In DAHL, we use gpt-4o from OpenAI\footnote{https://openai.com/} as the \textit{Splitter} with the prompt shown in Figure \ref{fig:prompt3}.  To assess whether gpt-4o can appropriately split each response into atomic units, we conducted statistical tests comparing the number of atomic units separated by gpt-4o and by our human annotator. Among 99 datasets, the T-test result had a p-value greater than 0.05, indicating no significant difference between the datasets and demonstrating that gpt-4o is a reliable splitter.\\

\textbf{Step2. Factuality Checking}\\ To verify the factuality of atomic units identified by the \textit{Splitter}, a reliable \textit{Checker} model is necessitated. As the judgments made by the \textit{Checker} must be credible, employing conventional LLMs  that rely solely on pretrained knowledge is insufficient. Therefore, for robust assessments, we utilize the pplx-API\footnote{https://www.perplexity.ai/} as our \textit{Checker}. Pplx-API draws upon information sourced from online documents to generate responses, thereby enhancing its credibility. Llama-3-8b-instruct \citep{dubey2024llama3herdmodels}  serves as the base LLM. We provide a single atomic unit and instruct the \textit{Checker} to determine its truthfulness in a binary form with the prompt in Figure \ref{fig:prompt4}. \\

\textbf{Step3. Score Calculation}\\ 
Responses with discrepancies in the number of atomic units and factuality labels are filtered out. The ratio of factual atomic units to total atomic units is calculated for each response, and the average score across all responses yields the final score. This final score is denoted as the \textit{DAHL Score}.

\section{Experiments}
Eight distinct models, including gpt-4o from OpenAI, Llama-3 and Llama-3.1 \citep{dubey2024llama3herdmodels}, Gemma-2 \citep{gemmateam2024gemma2improvingopen}, Qwen-2 \citep{yang2024qwen2technicalreport}, Mistral-Nemo-Base-2407 \footnote{https://huggingface.co/mistralai/Mistral-Nemo-Base-2407}, Dolly \citep{DatabricksBlog2023DollyV2}, Mpt \footnote{https://www.databricks.com/blog/mpt-7b} are experimented. As \citet{min2022rethinking} and \citet{webson2022promptbased} show, prompt templates or demonstrations tend to be helpful for task comprehension rather than directly enhancing performance. Therefore, our primary objective being hallucination evaluation focusing more on the factual precision than the format of response, we prompt the models to generate response from questions without additional prompt templates or demonstrations. Further, as we aim to evaluate hallucination rather than the instruction-following ability, we use pretrained models instead of fine-tuned models for evaluation. Response generation was done with the exact same generation configuration throughout the models for fair comparison, with temperature of 0.6 and maximum token length of 256.

\section{Results and Analysis}

\subsection{DAHL Score}

\begin{table}[]
\centering
\resizebox{\columnwidth}{!}{%
\renewcommand{\arraystretch}{1.3}
\begin{tabular}{c|cccc}
\toprule[1.3pt]
\textbf{Open}       & \textbf{Model Family}                    & \textbf{Size} & \textbf{Avg. Length} & \textbf{DAHL Score} \\ \toprule[1.3pt]
X                   & \textbf{gpt-4o}                   & ?             & 2321                 & \textbf{0.9365}     \\ \hline
\multirow{12}{*}{O} & \textbf{Llama-3}                  & 8B            & 1115                 & 0.8638              \\ \cline{2-5} 
                    & \multirow{2}{*}{\textbf{Llama-3.1}}        & 8B            & 1152                 & 0.8627              \\
                    &                                   & 70B           & 1156                    & 0.8733                   \\ \cline{2-5} 
                    & \multirow{2}{*}{\textbf{Gemma-2}} & 2B            & 1145                 & 0.8398              \\
                    &                                   & 9B            & 1145                 & 0.8765              \\ \cline{2-5} 
                    & \multirow{4}{*}{\textbf{Qwen-2}}  & 0.5B          & 1076                 & 0.7610              \\
                    &                                   & 1.5B          & 1067                 & 0.8151              \\
                    &                                   & 7B            & 1090                 & 0.8870     \\
                    &                                   & 72B           & 1062                    & \textbf{0.8997}                   \\ \cline{2-5} 
                    & \textbf{Mistral-Nemo-Base-2407}   & 12B           & 1127                 & 0.8087              \\ \cline{2-5} 
                    & \textbf{Dolly-v2}                 & 3B            & 1166                 & 0.8250              \\ \cline{2-5} 
                    & \textbf{Mpt}                      & 7B            & 1014                 & 0.7355              \\ \toprule[1.3pt]
\end{tabular}%
}
\caption{\textit{DAHL Score} and average length of responses(string) generated  from every model tested. Gpt-4o outperforms all, followed by Qwen-2, Gemma-2, Llama-3, Llama-3.1, Dolly-v2, Mistral-Nemo-Base-2407, and MPT. \textit{DAHL Score} of Qwen-2 with 72B parameters, the model with the highest score among the open-source models, along with that of gpt-4o, is marked bold.}
\label{tab:score_all}
\end{table}

As illustrated in Figure \ref{fig:pipeline}, we calculate the factual accuracy of each response utilizing the notion of Factscore \citep{min2023factscore}. The mean accuracy of responses across all questions is denoted as the \textit{DAHL Score} throughout this paper. The \textit{DAHL Score} of every model tested are listed in Table \ref{tab:score_all}. The categorical \textit{DAHL Score} of each model is provided in Figure \ref{fig:cat}. Gpt-4o outperforms all the models tested. Among the top performers in each model family, Qwen-2 has the highest \textit{DAHL Score} among open-source models, followed by Gemma-2, Llama-3.1, Llama-3, Dolly-v2, Mistral-Nemo-Base-2407, and Mpt.

The results in Table \ref{tab:score_all} indicate that \textbf{larger models tend to hallucinate less compared to smaller models in biomedicine.} Specifically, Llama-3.1 with 70 billion parameters has a higher \textit{DAHL Score} than its 8 billion parameter counterpart. Similarly, Gemma-2 demonstrates a significantly higher \textit{DAHL Score} in the 9 billion parameter model compared to the 2 billion parameter model, with a more pronounced difference than that observed between the two Llama-3.1 models. Qwen-2 also shows a linear increase in \textit{DAHL Score} with increasing model size, with the 72 billion parameter model achieving the highest score.

Interestingly, the difference in \textit{DAHL Score} between smaller models (such as the 2 billion and 9 billion parameter models of Gemma-2, or the 1.5 billion and 7 billion models of Qwen-2) is substantial, while the difference between larger models (such as Llama-3.1's 8 billion and 70 billion models, or Qwen-2's 7 billion and 72 billion models) is less pronounced. This suggests that \textbf{once a model reaches a certain size(7 to 8 billion), further scaling in model size lead to less prominent increase in factual accuracy of responses.}

\subsection{Effect of Temperature}

\begin{figure*}[t] 
    \centering
    \includegraphics[width=0.8\textwidth]{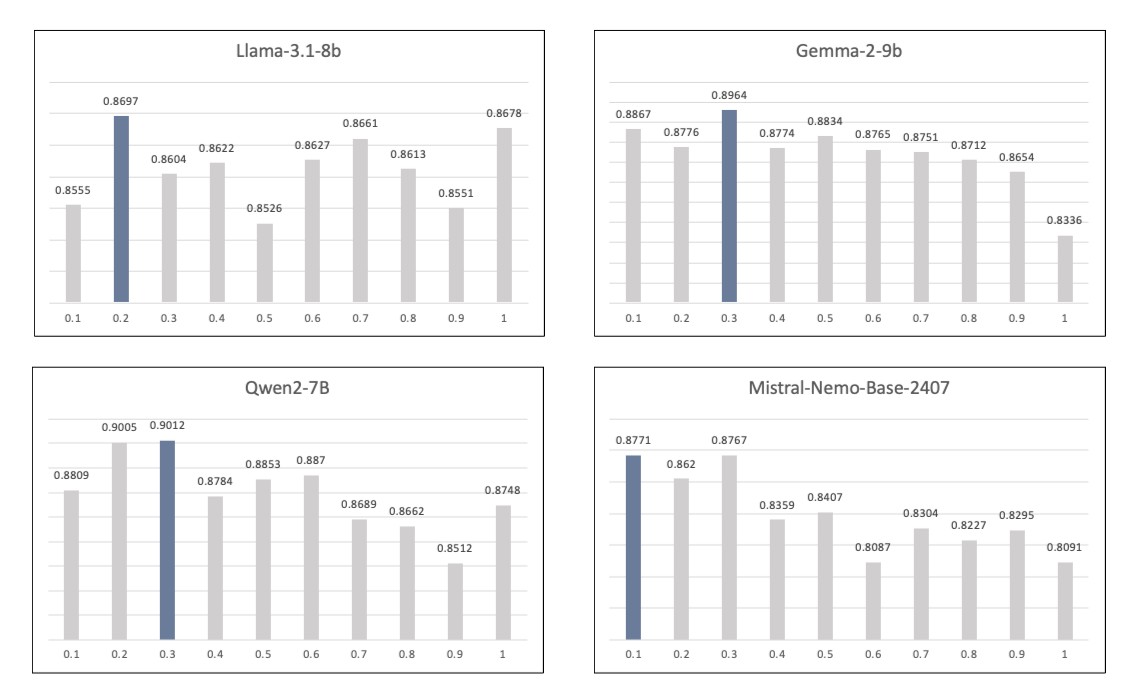} 
    \caption{The \textit{DAHL Score} for Llama-3.1-8b, Gemma-2-9b, Qwen-2-8b, and Mistral-Nemo-Base-2407 (12 billion parameters) evaluated across temperatures ranging from 0.1 to 1.0. The optimal temperature for each model falls within the range of 0.1 to 0.3, with a slight linear decrease in \textit{DAHL Score} as the temperature increases.}
    \label{fig:temp}
\end{figure*}

As \citet{xu2024hallucination} notes, hallucination is unavoidable. However, the ultimate goal of hallucination evaluation is its mitigation. Using DAHL, we conduct an ablation study with open-source base models to investigate whether adjusting the inference environment—specifically, the temperature setting—can reduce hallucination. To ensure a fair comparison, these experiments are performed on models of similar size. For efficiency, we use a 10\% sample of the question dataset, maintaining the same categorical distribution as the full dataset.

To validate the reliability of testing on the sampled data instead of the full dataset, we randomly select five different sample sets and evaluate the responses of gpt-4o for each. We then perform statistical comparisons on the resulting \textit{DAHL Scores}. Pairwise comparisons across the five sample sets yield a total of ten statistical tests. First, we conduct an F-test to assess the equality of variance between each pair of datasets. After confirming equal variances, we proceed with a T-test to determine if there are any significant differences between the sampled dataset pairs. All T-tests produce p-values (two-tailed) greater than 0.05, indicating no significant differences in the \textit{DAHL Score} across different sample sets. Therefore, we conclude that testing with 10\% of the sampled data sufficiently represents the evaluation scores obtained from the full dataset.

As depicted in Figure \ref{fig:temp}, we observe no consistent trend regarding temperature across four different models. There is no single optimal temperature universally applicable to all LLMs in terms of hallucination. However, \textbf{temperatures between 0.2 and 0.3 appear to provide the most suitable balance}, ensuring at least moderate performance across most models. We refrain from conducting further experiments with higher temperatures, as temperatures exceeding 1.0 are known to exacerbate hallucination \citep{renze2024effect}.

\subsection{Comparison with Human Evaluation}\label{human}

To validate the robustness of our automated hallucination evaluation methodology, we conducted a comparative analysis between the \textit{DAHL Score} and human evaluations of responses generated by LLMs. A set of 99 responses sampled with a balanced categorical distribution is utilized, ensuring that all 29 categories were represented for evaluation. Two human annotators independently assessed the factual precision of each response. Annotators were instructed to break down the responses into atomic units, each containing one piece of information. To ensure the reliability of the verification process, annotators were mandated to refer to trustworthy sources in the biomedical domain to judge the factuality of each atomic unit. Subsequently, they calculated the factual precision of each response by dividing the number of accurate facts by the total number of atomic units. Finally, we computed the average scores of the responses and compared them with the scores obtained from our automated process using various statistical tests. The Pearson Correlation Coefficient was found to be 0.5508 with a p-value of \(3.4927 \times 10^{-9}\), indicating a meaningful, moderate to strong correlation between automated and human evaluation scores.

Although there was a positive correlation between the human-annotated factuality scores and the \textit{DAHL Score}, we implemented a qualitative comparison between human annotations and our proposed automated hallucination evaluation pipeline using 33 randomly sampled responses, with one or two from each of the category. When comparing the atomic units identified by human annotators and our \textit{Splitter}, we observed that the results were generally similar; however, humans tended to split sentences involving coordinating conjunctions more frequently than the \textit{Splitter} does. For instance, given the sentence\\ 
{\itshape `The patient is completely \textcolor{gray}{unconscious} \textcolor{violet}{and} \textcolor{gray}{immobile}.'} \\
human annotators would split this into two atomic units: 

{\itshape `The patient is completely \textcolor{gray}{unconscious}.'}

{\itshape `The patient is completely \textcolor{gray}{immobile}.'}\\
In contrast, the \textit{Splitter} would treat this sentence as a single atomic unit. However, if a sentence contains coordinating conjunctions where the connected elements do not have equivalent relationships, the \textit{Splitter} tends to divide the sentence into multiple atomic units. For example, as Figure \ref{fig:splitter} shows, in the sentence \\
{\itshape `The most common locations for CL to occur are the neck and axilla (armpit), \textcolor{purple}{followed by} the \textcolor{gray}{mediastinum (chest)} \textcolor{violet}{and} \textcolor{gray}{retroperitoneum (abdomen)}.'} \\
the conjunction \textit{\textcolor{violet}{and}} connects \textcolor{gray}{\textit{mediastinum (chest)}} and \textcolor{gray}{\textit{retroperitoneum (abdomen)}}, but because the phrase \textcolor{purple}{\textit{followed by}} indicates an ordered sequence, the \textit{Splitter} splits the sentence into atomic units accordingly.

In the factuality-checking process, the main difference between human annotators and the \textit{Checker} is that human annotators were more likely to consider the context (i.e., the atomic units preceding or following the target unit) when assessing the factuality of an atomic unit. Even when it is difficult to evaluate the factuality of an atomic unit in isolation, such as when the subject includes a pronoun, human annotators may still make a judgment based on surrounding context. In contrast, the \textit{Checker} is more likely to label such cases as "false" or "unknown". If an atomic unit is labeled "unknown," the response containing that atomic unit is excluded from the final \textit{DAHL Score} calculation, as only responses definitively labeled as true or false are included. For instance, the \textit{Checker} might refrain from judging the factuality of a sentence like \textit{`\textcolor{gray}{This} patient is not aware of the procedure.'} This conservative approach indicates that the \textit{Checker} assesses the factuality of each atomic unit more stringently than human annotators.

\section{Conclusion and Future Work}

In this paper, we present DAHL, a benchmark dataset and an automated system designed to evaluate hallucination in long-form text generation within the biomedical domain. Utilizing a benchmark dataset consisting of 8,573 questions spanning 29 categories, our aim is to assess the hallucination with a specific focus on the factual precision of long-form responses. Diverging from previous methods that rely on accuracy calculations in classification or multiple-choice tasks, our approach involves deconstructing responses into atomic units and assessing the factuality of each unit. This nuanced methodology provides a more comprehensive evaluation of LLM hallucination. Our automatic dataset construction process ensures scalability across various domains and allows for regular updates. Moreover, our automated evaluation system offers significant efficiency compared to manual evaluation. Furthermore, the \textit{DAHL Score} generated by our system could potentially replace preference models, serving as labels for preference.

\section*{Limitation}

DAHL primarily targets fact-conflicting hallucination, yet it's essential to recognize the significance of input-conflicting and context-conflicting hallucination in evaluating LLMs. Additionally, the alignment of LLMs, specifically their ability to comprehend users' needs and generate helpful responses, should be considered for safe and satisfactory deployment of LLMs in real-life applications.

\section*{Ethics Statement}

While our proposed benchmark for evaluating the biomedical knowledge of LLMs is a step forward, it is important to note that it is not flawless. Consequently, the tested model with the best \textit{DAHL Score} should not be unquestioningly trusted or relied upon for clinical advice.

\bibliography{anthology,custom}

\begin{thebibliography}{36}
\expandafter\ifx\csname natexlab\endcsname\relax\def\natexlab#1{#1}\fi

\bibitem[{Cao et~al.(2023)Cao, Yang, and Zhao}]{cao2023autohall}
Zouying Cao, Yifei Yang, and Hai Zhao. 2023.
\newblock \href {http://arxiv.org/abs/2310.00259} {Autohall: Automated hallucination dataset generation for large language models}.

\bibitem[{Chen et~al.(2023)Chen, Zhao, Zhang, Chern, Gao, Liu, and He}]{chen2023felm}
Shiqi Chen, Yiran Zhao, Jinghan Zhang, I-Chun Chern, Siyang Gao, Pengfei Liu, and Junxian He. 2023.
\newblock \href {http://arxiv.org/abs/2310.00741} {Felm: Benchmarking factuality evaluation of large language models}.

\bibitem[{Conover et~al.(2023)Conover, Hayes, Mathur, Xie, Wan, Shah, Ghodsi, Wendell, Zaharia, and Xin}]{DatabricksBlog2023DollyV2}
Mike Conover, Matt Hayes, Ankit Mathur, Jianwei Xie, Jun Wan, Sam Shah, Ali Ghodsi, Patrick Wendell, Matei Zaharia, and Reynold Xin. 2023.
\newblock \href {https://www.databricks.com/blog/2023/04/12/dolly-first-open-commercially-viable-instruction-tuned-llm} {Free dolly: Introducing the world's first truly open instruction-tuned llm}.

\bibitem[{Dhuliawala et~al.(2023)Dhuliawala, Komeili, Xu, Raileanu, Li, Celikyilmaz, and Weston}]{dhuliawala2023chainofverification}
Shehzaad Dhuliawala, Mojtaba Komeili, Jing Xu, Roberta Raileanu, Xian Li, Asli Celikyilmaz, and Jason Weston. 2023.
\newblock \href {http://arxiv.org/abs/2309.11495} {Chain-of-verification reduces hallucination in large language models}.

\bibitem[{Dubey et~al.(2024)Dubey, Jauhri, Pandey, Kadian, Al-Dahle, Letman, Mathur, Schelten, Yang, Fan, Goyal, Hartshorn, Yang, Mitra, Sravankumar, Korenev, Hinsvark, Rao, Zhang, Rodriguez, Gregerson, Spataru, Roziere, Biron, Tang, Chern, Caucheteux, Nayak, Bi, Marra, McConnell, Keller, Touret, Wu, Wong, Ferrer, Nikolaidis, Allonsius, Song, Pintz, Livshits, Esiobu, Choudhary, Mahajan, Garcia-Olano, Perino, Hupkes, Lakomkin, AlBadawy, Lobanova, Dinan, Smith, Radenovic, Zhang, Synnaeve, Lee, Anderson, Nail, Mialon, Pang, Cucurell, Nguyen, Korevaar, Xu, Touvron, Zarov, Ibarra, Kloumann, Misra, Evtimov, Copet, Lee, Geffert, Vranes, Park, Mahadeokar, Shah, van~der Linde, Billock, Hong, Lee, Fu, Chi, Huang, Liu, Wang, Yu, Bitton, Spisak, Park, Rocca, Johnstun, Saxe, Jia, Alwala, Upasani, Plawiak, Li, Heafield, Stone, El-Arini, Iyer, Malik, Chiu, Bhalla, Rantala-Yeary, van~der Maaten, Chen, Tan, Jenkins, Martin, Madaan, Malo, Blecher, Landzaat, de~Oliveira, Muzzi, Pasupuleti, Singh, Paluri, Kardas, Oldham, Rita,
  Pavlova, Kambadur, Lewis, Si, Singh, Hassan, Goyal, Torabi, Bashlykov, Bogoychev, Chatterji, Duchenne, Çelebi, Alrassy, Zhang, Li, Vasic, Weng, Bhargava, Dubal, Krishnan, Koura, Xu, He, Dong, Srinivasan, Ganapathy, Calderer, Cabral, Stojnic, Raileanu, Girdhar, Patel, Sauvestre, Polidoro, Sumbaly, Taylor, Silva, Hou, Wang, Hosseini, Chennabasappa, Singh, Bell, Kim, Edunov, Nie, Narang, Raparthy, Shen, Wan, Bhosale, Zhang, Vandenhende, Batra, Whitman, Sootla, Collot, Gururangan, Borodinsky, Herman, Fowler, Sheasha, Georgiou, Scialom, Speckbacher, Mihaylov, Xiao, Karn, Goswami, Gupta, Ramanathan, Kerkez, Gonguet, Do, Vogeti, Petrovic, Chu, Xiong, Fu, Meers, Martinet, Wang, Tan, Xie, Jia, Wang, Goldschlag, Gaur, Babaei, Wen, Song, Zhang, Li, Mao, Coudert, Yan, Chen, Papakipos, Singh, Grattafiori, Jain, Kelsey, Shajnfeld, Gangidi, Victoria, Goldstand, Menon, Sharma, Boesenberg, Vaughan, Baevski, Feinstein, Kallet, Sangani, Yunus, Lupu, Alvarado, Caples, Gu, Ho, Poulton, Ryan, Ramchandani, Franco, Saraf,
  Chowdhury, Gabriel, Bharambe, Eisenman, Yazdan, James, Maurer, Leonhardi, Huang, Loyd, Paola, Paranjape, Liu, Wu, Ni, Hancock, Wasti, Spence, Stojkovic, Gamido, Montalvo, Parker, Burton, Mejia, Wang, Kim, Zhou, Hu, Chu, Cai, Tindal, Feichtenhofer, Civin, Beaty, Kreymer, Li, Wyatt, Adkins, Xu, Testuggine, David, Parikh, Liskovich, Foss, Wang, Le, Holland, Dowling, Jamil, Montgomery, Presani, Hahn, Wood, Brinkman, Arcaute, Dunbar, Smothers, Sun, Kreuk, Tian, Ozgenel, Caggioni, Guzmán, Kanayet, Seide, Florez, Schwarz, Badeer, Swee, Halpern, Thattai, Herman, Sizov, Guangyi, Zhang, Lakshminarayanan, Shojanazeri, Zou, Wang, Zha, Habeeb, Rudolph, Suk, Aspegren, Goldman, Molybog, Tufanov, Veliche, Gat, Weissman, Geboski, Kohli, Asher, Gaya, Marcus, Tang, Chan, Zhen, Reizenstein, Teboul, Zhong, Jin, Yang, Cummings, Carvill, Shepard, McPhie, Torres, Ginsburg, Wang, Wu, U, Saxena, Prasad, Khandelwal, Zand, Matosich, Veeraraghavan, Michelena, Li, Huang, Chawla, Lakhotia, Huang, Chen, Garg, A, Silva, Bell, Zhang, Guo,
  Yu, Moshkovich, Wehrstedt, Khabsa, Avalani, Bhatt, Tsimpoukelli, Mankus, Hasson, Lennie, Reso, Groshev, Naumov, Lathi, Keneally, Seltzer, Valko, Restrepo, Patel, Vyatskov, Samvelyan, Clark, Macey, Wang, Hermoso, Metanat, Rastegari, Bansal, Santhanam, Parks, White, Bawa, Singhal, Egebo, Usunier, Laptev, Dong, Zhang, Cheng, Chernoguz, Hart, Salpekar, Kalinli, Kent, Parekh, Saab, Balaji, Rittner, Bontrager, Roux, Dollar, Zvyagina, Ratanchandani, Yuvraj, Liang, Alao, Rodriguez, Ayub, Murthy, Nayani, Mitra, Li, Hogan, Battey, Wang, Maheswari, Howes, Rinott, Bondu, Datta, Chugh, Hunt, Dhillon, Sidorov, Pan, Verma, Yamamoto, Ramaswamy, Lindsay, Lindsay, Feng, Lin, Zha, Shankar, Zhang, Zhang, Wang, Agarwal, Sajuyigbe, Chintala, Max, Chen, Kehoe, Satterfield, Govindaprasad, Gupta, Cho, Virk, Subramanian, Choudhury, Goldman, Remez, Glaser, Best, Kohler, Robinson, Li, Zhang, Matthews, Chou, Shaked, Vontimitta, Ajayi, Montanez, Mohan, Kumar, Mangla, Ionescu, Poenaru, Mihailescu, Ivanov, Li, Wang, Jiang, Bouaziz,
  Constable, Tang, Wang, Wu, Wang, Xia, Wu, Gao, Chen, Hu, Jia, Qi, Li, Zhang, Zhang, Adi, Nam, Yu, Wang, Hao, Qian, He, Rait, DeVito, Rosnbrick, Wen, Yang, and Zhao}]{dubey2024llama3herdmodels}
Abhimanyu Dubey, Abhinav Jauhri, Abhinav Pandey, Abhishek Kadian, Ahmad Al-Dahle, Aiesha Letman, Akhil Mathur, Alan Schelten, Amy Yang, Angela Fan, Anirudh Goyal, Anthony Hartshorn, Aobo Yang, Archi Mitra, Archie Sravankumar, Artem Korenev, Arthur Hinsvark, Arun Rao, Aston Zhang, Aurelien Rodriguez, Austen Gregerson, Ava Spataru, Baptiste Roziere, Bethany Biron, Binh Tang, Bobbie Chern, Charlotte Caucheteux, Chaya Nayak, Chloe Bi, Chris Marra, Chris McConnell, Christian Keller, Christophe Touret, Chunyang Wu, Corinne Wong, Cristian~Canton Ferrer, Cyrus Nikolaidis, Damien Allonsius, Daniel Song, Danielle Pintz, Danny Livshits, David Esiobu, Dhruv Choudhary, Dhruv Mahajan, Diego Garcia-Olano, Diego Perino, Dieuwke Hupkes, Egor Lakomkin, Ehab AlBadawy, Elina Lobanova, Emily Dinan, Eric~Michael Smith, Filip Radenovic, Frank Zhang, Gabriel Synnaeve, Gabrielle Lee, Georgia~Lewis Anderson, Graeme Nail, Gregoire Mialon, Guan Pang, Guillem Cucurell, Hailey Nguyen, Hannah Korevaar, Hu~Xu, Hugo Touvron, Iliyan Zarov,
  Imanol~Arrieta Ibarra, Isabel Kloumann, Ishan Misra, Ivan Evtimov, Jade Copet, Jaewon Lee, Jan Geffert, Jana Vranes, Jason Park, Jay Mahadeokar, Jeet Shah, Jelmer van~der Linde, Jennifer Billock, Jenny Hong, Jenya Lee, Jeremy Fu, Jianfeng Chi, Jianyu Huang, Jiawen Liu, Jie Wang, Jiecao Yu, Joanna Bitton, Joe Spisak, Jongsoo Park, Joseph Rocca, Joshua Johnstun, Joshua Saxe, Junteng Jia, Kalyan~Vasuden Alwala, Kartikeya Upasani, Kate Plawiak, Ke~Li, Kenneth Heafield, Kevin Stone, Khalid El-Arini, Krithika Iyer, Kshitiz Malik, Kuenley Chiu, Kunal Bhalla, Lauren Rantala-Yeary, Laurens van~der Maaten, Lawrence Chen, Liang Tan, Liz Jenkins, Louis Martin, Lovish Madaan, Lubo Malo, Lukas Blecher, Lukas Landzaat, Luke de~Oliveira, Madeline Muzzi, Mahesh Pasupuleti, Mannat Singh, Manohar Paluri, Marcin Kardas, Mathew Oldham, Mathieu Rita, Maya Pavlova, Melanie Kambadur, Mike Lewis, Min Si, Mitesh~Kumar Singh, Mona Hassan, Naman Goyal, Narjes Torabi, Nikolay Bashlykov, Nikolay Bogoychev, Niladri Chatterji, Olivier
  Duchenne, Onur Çelebi, Patrick Alrassy, Pengchuan Zhang, Pengwei Li, Petar Vasic, Peter Weng, Prajjwal Bhargava, Pratik Dubal, Praveen Krishnan, Punit~Singh Koura, Puxin Xu, Qing He, Qingxiao Dong, Ragavan Srinivasan, Raj Ganapathy, Ramon Calderer, Ricardo~Silveira Cabral, Robert Stojnic, Roberta Raileanu, Rohit Girdhar, Rohit Patel, Romain Sauvestre, Ronnie Polidoro, Roshan Sumbaly, Ross Taylor, Ruan Silva, Rui Hou, Rui Wang, Saghar Hosseini, Sahana Chennabasappa, Sanjay Singh, Sean Bell, Seohyun~Sonia Kim, Sergey Edunov, Shaoliang Nie, Sharan Narang, Sharath Raparthy, Sheng Shen, Shengye Wan, Shruti Bhosale, Shun Zhang, Simon Vandenhende, Soumya Batra, Spencer Whitman, Sten Sootla, Stephane Collot, Suchin Gururangan, Sydney Borodinsky, Tamar Herman, Tara Fowler, Tarek Sheasha, Thomas Georgiou, Thomas Scialom, Tobias Speckbacher, Todor Mihaylov, Tong Xiao, Ujjwal Karn, Vedanuj Goswami, Vibhor Gupta, Vignesh Ramanathan, Viktor Kerkez, Vincent Gonguet, Virginie Do, Vish Vogeti, Vladan Petrovic, Weiwei Chu,
  Wenhan Xiong, Wenyin Fu, Whitney Meers, Xavier Martinet, Xiaodong Wang, Xiaoqing~Ellen Tan, Xinfeng Xie, Xuchao Jia, Xuewei Wang, Yaelle Goldschlag, Yashesh Gaur, Yasmine Babaei, Yi~Wen, Yiwen Song, Yuchen Zhang, Yue Li, Yuning Mao, Zacharie~Delpierre Coudert, Zheng Yan, Zhengxing Chen, Zoe Papakipos, Aaditya Singh, Aaron Grattafiori, Abha Jain, Adam Kelsey, Adam Shajnfeld, Adithya Gangidi, Adolfo Victoria, Ahuva Goldstand, Ajay Menon, Ajay Sharma, Alex Boesenberg, Alex Vaughan, Alexei Baevski, Allie Feinstein, Amanda Kallet, Amit Sangani, Anam Yunus, Andrei Lupu, Andres Alvarado, Andrew Caples, Andrew Gu, Andrew Ho, Andrew Poulton, Andrew Ryan, Ankit Ramchandani, Annie Franco, Aparajita Saraf, Arkabandhu Chowdhury, Ashley Gabriel, Ashwin Bharambe, Assaf Eisenman, Azadeh Yazdan, Beau James, Ben Maurer, Benjamin Leonhardi, Bernie Huang, Beth Loyd, Beto~De Paola, Bhargavi Paranjape, Bing Liu, Bo~Wu, Boyu Ni, Braden Hancock, Bram Wasti, Brandon Spence, Brani Stojkovic, Brian Gamido, Britt Montalvo, Carl
  Parker, Carly Burton, Catalina Mejia, Changhan Wang, Changkyu Kim, Chao Zhou, Chester Hu, Ching-Hsiang Chu, Chris Cai, Chris Tindal, Christoph Feichtenhofer, Damon Civin, Dana Beaty, Daniel Kreymer, Daniel Li, Danny Wyatt, David Adkins, David Xu, Davide Testuggine, Delia David, Devi Parikh, Diana Liskovich, Didem Foss, Dingkang Wang, Duc Le, Dustin Holland, Edward Dowling, Eissa Jamil, Elaine Montgomery, Eleonora Presani, Emily Hahn, Emily Wood, Erik Brinkman, Esteban Arcaute, Evan Dunbar, Evan Smothers, Fei Sun, Felix Kreuk, Feng Tian, Firat Ozgenel, Francesco Caggioni, Francisco Guzmán, Frank Kanayet, Frank Seide, Gabriela~Medina Florez, Gabriella Schwarz, Gada Badeer, Georgia Swee, Gil Halpern, Govind Thattai, Grant Herman, Grigory Sizov, Guangyi, Zhang, Guna Lakshminarayanan, Hamid Shojanazeri, Han Zou, Hannah Wang, Hanwen Zha, Haroun Habeeb, Harrison Rudolph, Helen Suk, Henry Aspegren, Hunter Goldman, Igor Molybog, Igor Tufanov, Irina-Elena Veliche, Itai Gat, Jake Weissman, James Geboski, James Kohli,
  Japhet Asher, Jean-Baptiste Gaya, Jeff Marcus, Jeff Tang, Jennifer Chan, Jenny Zhen, Jeremy Reizenstein, Jeremy Teboul, Jessica Zhong, Jian Jin, Jingyi Yang, Joe Cummings, Jon Carvill, Jon Shepard, Jonathan McPhie, Jonathan Torres, Josh Ginsburg, Junjie Wang, Kai Wu, Kam~Hou U, Karan Saxena, Karthik Prasad, Kartikay Khandelwal, Katayoun Zand, Kathy Matosich, Kaushik Veeraraghavan, Kelly Michelena, Keqian Li, Kun Huang, Kunal Chawla, Kushal Lakhotia, Kyle Huang, Lailin Chen, Lakshya Garg, Lavender A, Leandro Silva, Lee Bell, Lei Zhang, Liangpeng Guo, Licheng Yu, Liron Moshkovich, Luca Wehrstedt, Madian Khabsa, Manav Avalani, Manish Bhatt, Maria Tsimpoukelli, Martynas Mankus, Matan Hasson, Matthew Lennie, Matthias Reso, Maxim Groshev, Maxim Naumov, Maya Lathi, Meghan Keneally, Michael~L. Seltzer, Michal Valko, Michelle Restrepo, Mihir Patel, Mik Vyatskov, Mikayel Samvelyan, Mike Clark, Mike Macey, Mike Wang, Miquel~Jubert Hermoso, Mo~Metanat, Mohammad Rastegari, Munish Bansal, Nandhini Santhanam, Natascha
  Parks, Natasha White, Navyata Bawa, Nayan Singhal, Nick Egebo, Nicolas Usunier, Nikolay~Pavlovich Laptev, Ning Dong, Ning Zhang, Norman Cheng, Oleg Chernoguz, Olivia Hart, Omkar Salpekar, Ozlem Kalinli, Parkin Kent, Parth Parekh, Paul Saab, Pavan Balaji, Pedro Rittner, Philip Bontrager, Pierre Roux, Piotr Dollar, Polina Zvyagina, Prashant Ratanchandani, Pritish Yuvraj, Qian Liang, Rachad Alao, Rachel Rodriguez, Rafi Ayub, Raghotham Murthy, Raghu Nayani, Rahul Mitra, Raymond Li, Rebekkah Hogan, Robin Battey, Rocky Wang, Rohan Maheswari, Russ Howes, Ruty Rinott, Sai~Jayesh Bondu, Samyak Datta, Sara Chugh, Sara Hunt, Sargun Dhillon, Sasha Sidorov, Satadru Pan, Saurabh Verma, Seiji Yamamoto, Sharadh Ramaswamy, Shaun Lindsay, Shaun Lindsay, Sheng Feng, Shenghao Lin, Shengxin~Cindy Zha, Shiva Shankar, Shuqiang Zhang, Shuqiang Zhang, Sinong Wang, Sneha Agarwal, Soji Sajuyigbe, Soumith Chintala, Stephanie Max, Stephen Chen, Steve Kehoe, Steve Satterfield, Sudarshan Govindaprasad, Sumit Gupta, Sungmin Cho, Sunny
  Virk, Suraj Subramanian, Sy~Choudhury, Sydney Goldman, Tal Remez, Tamar Glaser, Tamara Best, Thilo Kohler, Thomas Robinson, Tianhe Li, Tianjun Zhang, Tim Matthews, Timothy Chou, Tzook Shaked, Varun Vontimitta, Victoria Ajayi, Victoria Montanez, Vijai Mohan, Vinay~Satish Kumar, Vishal Mangla, Vlad Ionescu, Vlad Poenaru, Vlad~Tiberiu Mihailescu, Vladimir Ivanov, Wei Li, Wenchen Wang, Wenwen Jiang, Wes Bouaziz, Will Constable, Xiaocheng Tang, Xiaofang Wang, Xiaojian Wu, Xiaolan Wang, Xide Xia, Xilun Wu, Xinbo Gao, Yanjun Chen, Ye~Hu, Ye~Jia, Ye~Qi, Yenda Li, Yilin Zhang, Ying Zhang, Yossi Adi, Youngjin Nam, Yu, Wang, Yuchen Hao, Yundi Qian, Yuzi He, Zach Rait, Zachary DeVito, Zef Rosnbrick, Zhaoduo Wen, Zhenyu Yang, and Zhiwei Zhao. 2024.
\newblock \href {http://arxiv.org/abs/2407.21783} {The llama 3 herd of models}.

\bibitem[{Durmus et~al.(2020)Durmus, He, and Diab}]{Durmus_2020}
Esin Durmus, He~He, and Mona Diab. 2020.
\newblock \href {https://doi.org/10.18653/v1/2020.acl-main.454} {Feqa: A question answering evaluation framework for faithfulness assessment in abstractive summarization}.
\newblock In \emph{Proceedings of the 58th Annual Meeting of the Association for Computational Linguistics}. Association for Computational Linguistics.

\bibitem[{Elaraby et~al.(2023)Elaraby, Lu, Dunn, Zhang, Wang, Liu, Tian, Wang, and Wang}]{elaraby2023halo}
Mohamed Elaraby, Mengyin Lu, Jacob Dunn, Xueying Zhang, Yu~Wang, Shizhu Liu, Pingchuan Tian, Yuping Wang, and Yuxuan Wang. 2023.
\newblock \href {http://arxiv.org/abs/2308.11764} {Halo: Estimation and reduction of hallucinations in open-source weak large language models}.

\bibitem[{Goodrich et~al.(2019)Goodrich, Rao, Liu, and Saleh}]{Goodrich_2019}
Ben Goodrich, Vinay Rao, Peter~J. Liu, and Mohammad Saleh. 2019.
\newblock \href {https://doi.org/10.1145/3292500.3330955} {Assessing the factual accuracy of generated text}.
\newblock In \emph{Proceedings of the 25th ACM SIGKDD International Conference on Knowledge Discovery \&; Data Mining}, KDD ’19. ACM.

\bibitem[{Gu et~al.(2023)Gu, Zhu, Ye, Zhang, Wang, Jiang, Xiong, Li, He, Xu, Huang, Wang, Wang, Zheng, Feng, and Xiao}]{gu2023xiezhi}
Zhouhong Gu, Xiaoxuan Zhu, Haoning Ye, Lin Zhang, Jianchen Wang, Sihang Jiang, Zhuozhi Xiong, Zihan Li, Qianyu He, Rui Xu, Wenhao Huang, Zili Wang, Shusen Wang, Weiguo Zheng, Hongwei Feng, and Yanghua Xiao. 2023.
\newblock \href {http://arxiv.org/abs/2306.05783} {Xiezhi: An ever-updating benchmark for holistic domain knowledge evaluation}.

\bibitem[{Ji et~al.(2023)Ji, Lee, Frieske, Yu, Su, Xu, Ishii, Bang, Madotto, and Fung}]{Ji_2023}
Ziwei Ji, Nayeon Lee, Rita Frieske, Tiezheng Yu, Dan Su, Yan Xu, Etsuko Ishii, Ye~Jin Bang, Andrea Madotto, and Pascale Fung. 2023.
\newblock \href {https://doi.org/10.1145/3571730} {Survey of hallucination in natural language generation}.
\newblock \emph{ACM Computing Surveys}, 55(12).

\bibitem[{Kadavath et~al.(2022)Kadavath, Conerly, Askell, Henighan, Drain, Perez, Schiefer, Hatfield-Dodds, DasSarma, Tran-Johnson, Johnston, El-Showk, Jones, Elhage, Hume, Chen, Bai, Bowman, Fort, Ganguli, Hernandez, Jacobson, Kernion, Kravec, Lovitt, Ndousse, Olsson, Ringer, Amodei, Brown, Clark, Joseph, Mann, McCandlish, Olah, and Kaplan}]{kadavath2022language}
Saurav Kadavath, Tom Conerly, Amanda Askell, Tom Henighan, Dawn Drain, Ethan Perez, Nicholas Schiefer, Zac Hatfield-Dodds, Nova DasSarma, Eli Tran-Johnson, Scott Johnston, Sheer El-Showk, Andy Jones, Nelson Elhage, Tristan Hume, Anna Chen, Yuntao Bai, Sam Bowman, Stanislav Fort, Deep Ganguli, Danny Hernandez, Josh Jacobson, Jackson Kernion, Shauna Kravec, Liane Lovitt, Kamal Ndousse, Catherine Olsson, Sam Ringer, Dario Amodei, Tom Brown, Jack Clark, Nicholas Joseph, Ben Mann, Sam McCandlish, Chris Olah, and Jared Kaplan. 2022.
\newblock \href {http://arxiv.org/abs/2207.05221} {Language models (mostly) know what they know}.

\bibitem[{Kandpal et~al.(2023)Kandpal, Deng, Roberts, Wallace, and Raffel}]{kandpal2023large}
Nikhil Kandpal, Haikang Deng, Adam Roberts, Eric Wallace, and Colin Raffel. 2023.
\newblock \href {http://arxiv.org/abs/2211.08411} {Large language models struggle to learn long-tail knowledge}.

\bibitem[{Lakkaraju et~al.(2022)Lakkaraju, Slack, Chen, Tan, and Singh}]{lakkaraju2022rethinking}
Himabindu Lakkaraju, Dylan Slack, Yuxin Chen, Chenhao Tan, and Sameer Singh. 2022.
\newblock \href {http://arxiv.org/abs/2202.01875} {Rethinking explainability as a dialogue: A practitioner's perspective}.

\bibitem[{Lee et~al.(2023)Lee, Bubeck, and Petro}]{article}
Peter Lee, Sebastien Bubeck, and Joseph Petro. 2023.
\newblock \href {https://doi.org/10.1056/NEJMsr2214184} {Benefits, limits, and risks of gpt-4 as an ai chatbot for medicine}.
\newblock \emph{The New England journal of medicine}, 388:1233--1239.

\bibitem[{Li et~al.(2024)Li, Chen, Ren, Cheng, Zhao, Nie, and Wen}]{li2024dawn}
Junyi Li, Jie Chen, Ruiyang Ren, Xiaoxue Cheng, Wayne~Xin Zhao, Jian-Yun Nie, and Ji-Rong Wen. 2024.
\newblock \href {http://arxiv.org/abs/2401.03205} {The dawn after the dark: An empirical study on factuality hallucination in large language models}.

\bibitem[{Li et~al.(2023)Li, Li, Zhang, Dan, Jiang, and Zhang}]{li2023chatdoctor}
Yunxiang Li, Zihan Li, Kai Zhang, Ruilong Dan, Steve Jiang, and You Zhang. 2023.
\newblock \href {http://arxiv.org/abs/2303.14070} {Chatdoctor: A medical chat model fine-tuned on a large language model meta-ai (llama) using medical domain knowledge}.

\bibitem[{Liao et~al.(2023)Liao, Meng, Liu, Wang, and Wang}]{liao2023automatic}
Yusheng Liao, Yutong Meng, Hongcheng Liu, Yanfeng Wang, and Yu~Wang. 2023.
\newblock \href {http://arxiv.org/abs/2309.02077} {An automatic evaluation framework for multi-turn medical consultations capabilities of large language models}.

\bibitem[{Lin et~al.(2022)Lin, Hilton, and Evans}]{lin2022truthfulqa}
Stephanie Lin, Jacob Hilton, and Owain Evans. 2022.
\newblock \href {http://arxiv.org/abs/2109.07958} {Truthfulqa: Measuring how models mimic human falsehoods}.

\bibitem[{Mihaylova et~al.(2019)Mihaylova, Karadjov, Atanasova, Baly, Mohtarami, and Nakov}]{mihaylova2019semeval2019}
Tsvetomila Mihaylova, Georgi Karadjov, Pepa Atanasova, Ramy Baly, Mitra Mohtarami, and Preslav Nakov. 2019.
\newblock \href {http://arxiv.org/abs/1906.01727} {Semeval-2019 task 8: Fact checking in community question answering forums}.

\bibitem[{Min et~al.(2023)Min, Krishna, Lyu, Lewis, tau Yih, Koh, Iyyer, Zettlemoyer, and Hajishirzi}]{min2023factscore}
Sewon Min, Kalpesh Krishna, Xinxi Lyu, Mike Lewis, Wen tau Yih, Pang~Wei Koh, Mohit Iyyer, Luke Zettlemoyer, and Hannaneh Hajishirzi. 2023.
\newblock \href {http://arxiv.org/abs/2305.14251} {Factscore: Fine-grained atomic evaluation of factual precision in long form text generation}.

\bibitem[{Min et~al.(2022)Min, Lyu, Holtzman, Artetxe, Lewis, Hajishirzi, and Zettlemoyer}]{min2022rethinking}
Sewon Min, Xinxi Lyu, Ari Holtzman, Mikel Artetxe, Mike Lewis, Hannaneh Hajishirzi, and Luke Zettlemoyer. 2022.
\newblock \href {http://arxiv.org/abs/2202.12837} {Rethinking the role of demonstrations: What makes in-context learning work?}

\bibitem[{Pal et~al.(2022)Pal, Umapathi, and Sankarasubbu}]{pmlr-v174-pal22a}
Ankit Pal, Logesh~Kumar Umapathi, and Malaikannan Sankarasubbu. 2022.
\newblock \href {https://proceedings.mlr.press/v174/pal22a.html} {Medmcqa: A large-scale multi-subject multi-choice dataset for medical domain question answering}.
\newblock In \emph{Proceedings of the Conference on Health, Inference, and Learning}, volume 174 of \emph{Proceedings of Machine Learning Research}, pages 248--260. PMLR.

\bibitem[{Pal et~al.(2023)Pal, Umapathi, and Sankarasubbu}]{pal2023medhalt}
Ankit Pal, Logesh~Kumar Umapathi, and Malaikannan Sankarasubbu. 2023.
\newblock \href {http://arxiv.org/abs/2307.15343} {Med-halt: Medical domain hallucination test for large language models}.

\bibitem[{Rawte et~al.(2023{\natexlab{a}})Rawte, Chakraborty, Pathak, Sarkar, Tonmoy, Chadha, Sheth, and Das}]{rawte2023troubling}
Vipula Rawte, Swagata Chakraborty, Agnibh Pathak, Anubhav Sarkar, S.~M Towhidul~Islam Tonmoy, Aman Chadha, Amit~P. Sheth, and Amitava Das. 2023{\natexlab{a}}.
\newblock \href {http://arxiv.org/abs/2310.04988} {The troubling emergence of hallucination in large language models -- an extensive definition, quantification, and prescriptive remediations}.

\bibitem[{Rawte et~al.(2023{\natexlab{b}})Rawte, Sheth, and Das}]{rawte2023survey}
Vipula Rawte, Amit Sheth, and Amitava Das. 2023{\natexlab{b}}.
\newblock \href {http://arxiv.org/abs/2309.05922} {A survey of hallucination in large foundation models}.

\bibitem[{Renze and Guven(2024)}]{renze2024effect}
Matthew Renze and Erhan Guven. 2024.
\newblock \href {http://arxiv.org/abs/2402.05201} {The effect of sampling temperature on problem solving in large language models}.

\bibitem[{Singhal et~al.(2022)Singhal, Azizi, Tu, Mahdavi, Wei, Chung, Scales, Tanwani, Cole-Lewis, Pfohl, Payne, Seneviratne, Gamble, Kelly, Scharli, Chowdhery, Mansfield, y~Arcas, Webster, Corrado, Matias, Chou, Gottweis, Tomasev, Liu, Rajkomar, Barral, Semturs, Karthikesalingam, and Natarajan}]{singhal2022large}
Karan Singhal, Shekoofeh Azizi, Tao Tu, S.~Sara Mahdavi, Jason Wei, Hyung~Won Chung, Nathan Scales, Ajay Tanwani, Heather Cole-Lewis, Stephen Pfohl, Perry Payne, Martin Seneviratne, Paul Gamble, Chris Kelly, Nathaneal Scharli, Aakanksha Chowdhery, Philip Mansfield, Blaise~Aguera y~Arcas, Dale Webster, Greg~S. Corrado, Yossi Matias, Katherine Chou, Juraj Gottweis, Nenad Tomasev, Yun Liu, Alvin Rajkomar, Joelle Barral, Christopher Semturs, Alan Karthikesalingam, and Vivek Natarajan. 2022.
\newblock \href {http://arxiv.org/abs/2212.13138} {Large language models encode clinical knowledge}.

\bibitem[{Team et~al.(2024)Team, Riviere, Pathak, Sessa, Hardin, Bhupatiraju, Hussenot, Mesnard, Shahriari, Ramé, Ferret, Liu, Tafti, Friesen, Casbon, Ramos, Kumar, Lan, Jerome, Tsitsulin, Vieillard, Stanczyk, Girgin, Momchev, Hoffman, Thakoor, Grill, Neyshabur, Bachem, Walton, Severyn, Parrish, Ahmad, Hutchison, Abdagic, Carl, Shen, Brock, Coenen, Laforge, Paterson, Bastian, Piot, Wu, Royal, Chen, Kumar, Perry, Welty, Choquette-Choo, Sinopalnikov, Weinberger, Vijaykumar, Rogozińska, Herbison, Bandy, Wang, Noland, Moreira, Senter, Eltyshev, Visin, Rasskin, Wei, Cameron, Martins, Hashemi, Klimczak-Plucińska, Batra, Dhand, Nardini, Mein, Zhou, Svensson, Stanway, Chan, Zhou, Carrasqueira, Iljazi, Becker, Fernandez, van Amersfoort, Gordon, Lipschultz, Newlan, yeong Ji, Mohamed, Badola, Black, Millican, McDonell, Nguyen, Sodhia, Greene, Sjoesund, Usui, Sifre, Heuermann, Lago, McNealus, Soares, Kilpatrick, Dixon, Martins, Reid, Singh, Iverson, Görner, Velloso, Wirth, Davidow, Miller, Rahtz, Watson, Risdal,
  Kazemi, Moynihan, Zhang, Kahng, Park, Rahman, Khatwani, Dao, Bardoliwalla, Devanathan, Dumai, Chauhan, Wahltinez, Botarda, Barnes, Barham, Michel, Jin, Georgiev, Culliton, Kuppala, Comanescu, Merhej, Jana, Rokni, Agarwal, Mullins, Saadat, Carthy, Perrin, Arnold, Krause, Dai, Garg, Sheth, Ronstrom, Chan, Jordan, Yu, Eccles, Hennigan, Kocisky, Doshi, Jain, Yadav, Meshram, Dharmadhikari, Barkley, Wei, Ye, Han, Kwon, Xu, Shen, Gong, Wei, Cotruta, Kirk, Rao, Giang, Peran, Warkentin, Collins, Barral, Ghahramani, Hadsell, Sculley, Banks, Dragan, Petrov, Vinyals, Dean, Hassabis, Kavukcuoglu, Farabet, Buchatskaya, Borgeaud, Fiedel, Joulin, Kenealy, Dadashi, and Andreev}]{gemmateam2024gemma2improvingopen}
Gemma Team, Morgane Riviere, Shreya Pathak, Pier~Giuseppe Sessa, Cassidy Hardin, Surya Bhupatiraju, Léonard Hussenot, Thomas Mesnard, Bobak Shahriari, Alexandre Ramé, Johan Ferret, Peter Liu, Pouya Tafti, Abe Friesen, Michelle Casbon, Sabela Ramos, Ravin Kumar, Charline~Le Lan, Sammy Jerome, Anton Tsitsulin, Nino Vieillard, Piotr Stanczyk, Sertan Girgin, Nikola Momchev, Matt Hoffman, Shantanu Thakoor, Jean-Bastien Grill, Behnam Neyshabur, Olivier Bachem, Alanna Walton, Aliaksei Severyn, Alicia Parrish, Aliya Ahmad, Allen Hutchison, Alvin Abdagic, Amanda Carl, Amy Shen, Andy Brock, Andy Coenen, Anthony Laforge, Antonia Paterson, Ben Bastian, Bilal Piot, Bo~Wu, Brandon Royal, Charlie Chen, Chintu Kumar, Chris Perry, Chris Welty, Christopher~A. Choquette-Choo, Danila Sinopalnikov, David Weinberger, Dimple Vijaykumar, Dominika Rogozińska, Dustin Herbison, Elisa Bandy, Emma Wang, Eric Noland, Erica Moreira, Evan Senter, Evgenii Eltyshev, Francesco Visin, Gabriel Rasskin, Gary Wei, Glenn Cameron, Gus Martins,
  Hadi Hashemi, Hanna Klimczak-Plucińska, Harleen Batra, Harsh Dhand, Ivan Nardini, Jacinda Mein, Jack Zhou, James Svensson, Jeff Stanway, Jetha Chan, Jin~Peng Zhou, Joana Carrasqueira, Joana Iljazi, Jocelyn Becker, Joe Fernandez, Joost van Amersfoort, Josh Gordon, Josh Lipschultz, Josh Newlan, Ju~yeong Ji, Kareem Mohamed, Kartikeya Badola, Kat Black, Katie Millican, Keelin McDonell, Kelvin Nguyen, Kiranbir Sodhia, Kish Greene, Lars~Lowe Sjoesund, Lauren Usui, Laurent Sifre, Lena Heuermann, Leticia Lago, Lilly McNealus, Livio~Baldini Soares, Logan Kilpatrick, Lucas Dixon, Luciano Martins, Machel Reid, Manvinder Singh, Mark Iverson, Martin Görner, Mat Velloso, Mateo Wirth, Matt Davidow, Matt Miller, Matthew Rahtz, Matthew Watson, Meg Risdal, Mehran Kazemi, Michael Moynihan, Ming Zhang, Minsuk Kahng, Minwoo Park, Mofi Rahman, Mohit Khatwani, Natalie Dao, Nenshad Bardoliwalla, Nesh Devanathan, Neta Dumai, Nilay Chauhan, Oscar Wahltinez, Pankil Botarda, Parker Barnes, Paul Barham, Paul Michel, Pengchong Jin,
  Petko Georgiev, Phil Culliton, Pradeep Kuppala, Ramona Comanescu, Ramona Merhej, Reena Jana, Reza~Ardeshir Rokni, Rishabh Agarwal, Ryan Mullins, Samaneh Saadat, Sara~Mc Carthy, Sarah Perrin, Sébastien M.~R. Arnold, Sebastian Krause, Shengyang Dai, Shruti Garg, Shruti Sheth, Sue Ronstrom, Susan Chan, Timothy Jordan, Ting Yu, Tom Eccles, Tom Hennigan, Tomas Kocisky, Tulsee Doshi, Vihan Jain, Vikas Yadav, Vilobh Meshram, Vishal Dharmadhikari, Warren Barkley, Wei Wei, Wenming Ye, Woohyun Han, Woosuk Kwon, Xiang Xu, Zhe Shen, Zhitao Gong, Zichuan Wei, Victor Cotruta, Phoebe Kirk, Anand Rao, Minh Giang, Ludovic Peran, Tris Warkentin, Eli Collins, Joelle Barral, Zoubin Ghahramani, Raia Hadsell, D.~Sculley, Jeanine Banks, Anca Dragan, Slav Petrov, Oriol Vinyals, Jeff Dean, Demis Hassabis, Koray Kavukcuoglu, Clement Farabet, Elena Buchatskaya, Sebastian Borgeaud, Noah Fiedel, Armand Joulin, Kathleen Kenealy, Robert Dadashi, and Alek Andreev. 2024.
\newblock \href {http://arxiv.org/abs/2408.00118} {Gemma 2: Improving open language models at a practical size}.

\bibitem[{Webson and Pavlick(2022)}]{webson2022promptbased}
Albert Webson and Ellie Pavlick. 2022.
\newblock \href {http://arxiv.org/abs/2109.01247} {Do prompt-based models really understand the meaning of their prompts?}

\bibitem[{Wei et~al.(2023)Wei, Wang, Schuurmans, Bosma, Ichter, Xia, Chi, Le, and Zhou}]{wei2023chainofthought}
Jason Wei, Xuezhi Wang, Dale Schuurmans, Maarten Bosma, Brian Ichter, Fei Xia, Ed~Chi, Quoc Le, and Denny Zhou. 2023.
\newblock \href {http://arxiv.org/abs/2201.11903} {Chain-of-thought prompting elicits reasoning in large language models}.

\bibitem[{Xu et~al.(2024)Xu, Jain, and Kankanhalli}]{xu2024hallucination}
Ziwei Xu, Sanjay Jain, and Mohan Kankanhalli. 2024.
\newblock \href {http://arxiv.org/abs/2401.11817} {Hallucination is inevitable: An innate limitation of large language models}.

\bibitem[{Yang et~al.(2024)Yang, Yang, Hui, Zheng, Yu, Zhou, Li, Li, Liu, Huang, Dong, Wei, Lin, Tang, Wang, Yang, Tu, Zhang, Ma, Yang, Xu, Zhou, Bai, He, Lin, Dang, Lu, Chen, Yang, Li, Xue, Ni, Zhang, Wang, Peng, Men, Gao, Lin, Wang, Bai, Tan, Zhu, Li, Liu, Ge, Deng, Zhou, Ren, Zhang, Wei, Ren, Liu, Fan, Yao, Zhang, Wan, Chu, Liu, Cui, Zhang, Guo, and Fan}]{yang2024qwen2technicalreport}
An~Yang, Baosong Yang, Binyuan Hui, Bo~Zheng, Bowen Yu, Chang Zhou, Chengpeng Li, Chengyuan Li, Dayiheng Liu, Fei Huang, Guanting Dong, Haoran Wei, Huan Lin, Jialong Tang, Jialin Wang, Jian Yang, Jianhong Tu, Jianwei Zhang, Jianxin Ma, Jianxin Yang, Jin Xu, Jingren Zhou, Jinze Bai, Jinzheng He, Junyang Lin, Kai Dang, Keming Lu, Keqin Chen, Kexin Yang, Mei Li, Mingfeng Xue, Na~Ni, Pei Zhang, Peng Wang, Ru~Peng, Rui Men, Ruize Gao, Runji Lin, Shijie Wang, Shuai Bai, Sinan Tan, Tianhang Zhu, Tianhao Li, Tianyu Liu, Wenbin Ge, Xiaodong Deng, Xiaohuan Zhou, Xingzhang Ren, Xinyu Zhang, Xipin Wei, Xuancheng Ren, Xuejing Liu, Yang Fan, Yang Yao, Yichang Zhang, Yu~Wan, Yunfei Chu, Yuqiong Liu, Zeyu Cui, Zhenru Zhang, Zhifang Guo, and Zhihao Fan. 2024.
\newblock \href {http://arxiv.org/abs/2407.10671} {Qwen2 technical report}.

\bibitem[{Yang et~al.(2023)Yang, Wang, Greenblatt, Huang, and Zhang}]{clinical_chatbot}
He~Yang, Fei Wang, Matthew Greenblatt, Sharon Huang, and Yi~Zhang. 2023.
\newblock \href {https://doi.org/10.1093/clinchem/hvad106} {Ai chatbots in clinical laboratory medicine: Foundations and trends}.
\newblock \emph{Clinical chemistry}, 69.

\bibitem[{Zhang et~al.(2023)Zhang, Li, Cui, Cai, Liu, Fu, Huang, Zhao, Zhang, Chen, Wang, Luu, Bi, Shi, and Shi}]{zhang2023sirens}
Yue Zhang, Yafu Li, Leyang Cui, Deng Cai, Lemao Liu, Tingchen Fu, Xinting Huang, Enbo Zhao, Yu~Zhang, Yulong Chen, Longyue Wang, Anh~Tuan Luu, Wei Bi, Freda Shi, and Shuming Shi. 2023.
\newblock \href {http://arxiv.org/abs/2309.01219} {Siren's song in the ai ocean: A survey on hallucination in large language models}.

\bibitem[{Zhao et~al.(2023)Zhao, Zhou, Li, Tang, Wang, Hou, Min, Zhang, Zhang, Dong, Du, Yang, Chen, Chen, Jiang, Ren, Li, Tang, Liu, Liu, Nie, and Wen}]{zhao2023survey}
Wayne~Xin Zhao, Kun Zhou, Junyi Li, Tianyi Tang, Xiaolei Wang, Yupeng Hou, Yingqian Min, Beichen Zhang, Junjie Zhang, Zican Dong, Yifan Du, Chen Yang, Yushuo Chen, Zhipeng Chen, Jinhao Jiang, Ruiyang Ren, Yifan Li, Xinyu Tang, Zikang Liu, Peiyu Liu, Jian-Yun Nie, and Ji-Rong Wen. 2023.
\newblock \href {http://arxiv.org/abs/2303.18223} {A survey of large language models}.

\bibitem[{Zhou et~al.(2024)Zhou, Liu, Gu, Zou, Huang, Wu, Li, Chen, Zhou, Liu, Hua, Mao, You, Wu, Zheng, Clifton, Li, Luo, and Clifton}]{zhou2024survey}
Hongjian Zhou, Fenglin Liu, Boyang Gu, Xinyu Zou, Jinfa Huang, Jinge Wu, Yiru Li, Sam~S. Chen, Peilin Zhou, Junling Liu, Yining Hua, Chengfeng Mao, Chenyu You, Xian Wu, Yefeng Zheng, Lei Clifton, Zheng Li, Jiebo Luo, and David~A. Clifton. 2024.
\newblock \href {http://arxiv.org/abs/2311.05112} {A survey of large language models in medicine: Principles, applications, and challenges}.

\end{thebibliography}
\bibliographystyle{acl_natbib}

\newpage

\appendix
\begin{appendices}

\section{Automatically Filtered Questions }
\label{sec:filtered}
\subsection{the/this/that/... + study/analysis/paper/... }
\begin{itemize}
    \item What ethical considerations are addressed by \textbf{the authors} in relation to \textbf{their research findings}?
    \item What are the implications for practice suggested by \textbf{the study}?
\end{itemize}

\subsection{mentioned/inferred/addressed/...}
\begin{itemize}
    \item What tissue-specific patterns were \textbf{observed} in the usage of intronic PASs compared to PASs in exons?
    \item What challenges are associated with the protocol described in the study, and what solutions are \textbf{suggested} for troubleshooting?

\end{itemize}

\subsection{was/were + used/... }
\begin{itemize}
    \item What method \textbf{was used} to assess the functional accuracy of the context-specific models?

\end{itemize}

\section{Manually Filtered Questions}

\label{sec:unfiltered}

Although the majority of questions containing the specified expressions are deleted, exceptions exist where questions can still be answered without additional context, even when they include these expressions. Consequently, questions containing the specified expressions are manually removed to enhance the quality of the benchmark dataset and to preserve as many questions as possible.

\subsection{based on/depending on/according to/... }
\begin{itemize}
    \item Explain the significance of functional validation \textbf{in the context of} this research and how it is achieved.
    
\end{itemize}
\subsection{was/were/did}
\begin{itemize}

    \item Which MEM \textbf{was} found to be the most computationally efficient, and how might this impact its use in research?
\end{itemize}

\begin{figure*}[h] 
    \centering
    \includegraphics[width=\textwidth]{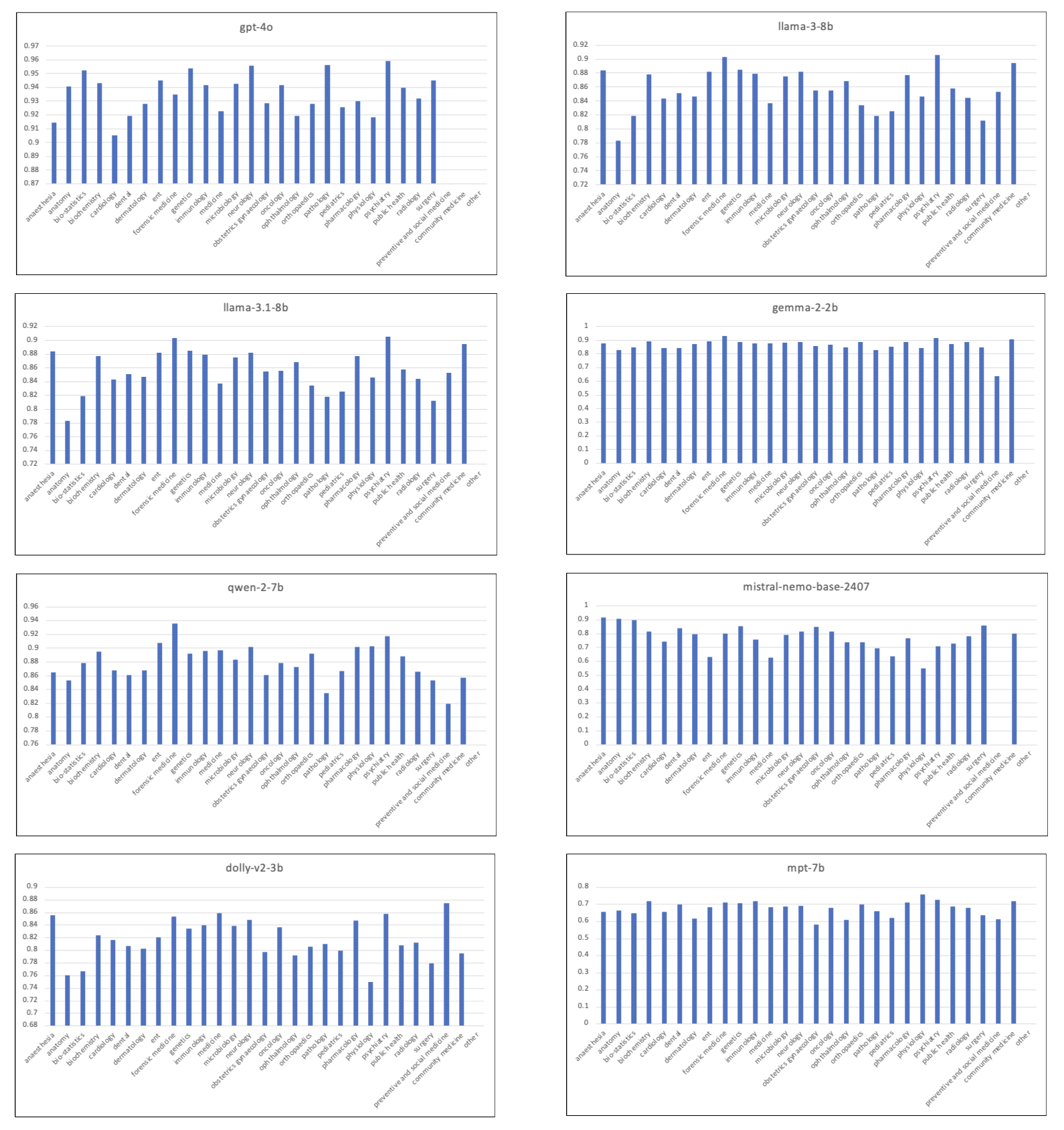} 
    \caption{Categorical \textit{DAHL Score} of each model.}
    \label{fig:cat}
\end{figure*}

\newpage

\begin{figure*}[ht] 
    \centering
    \includegraphics[width=0.6\textwidth]{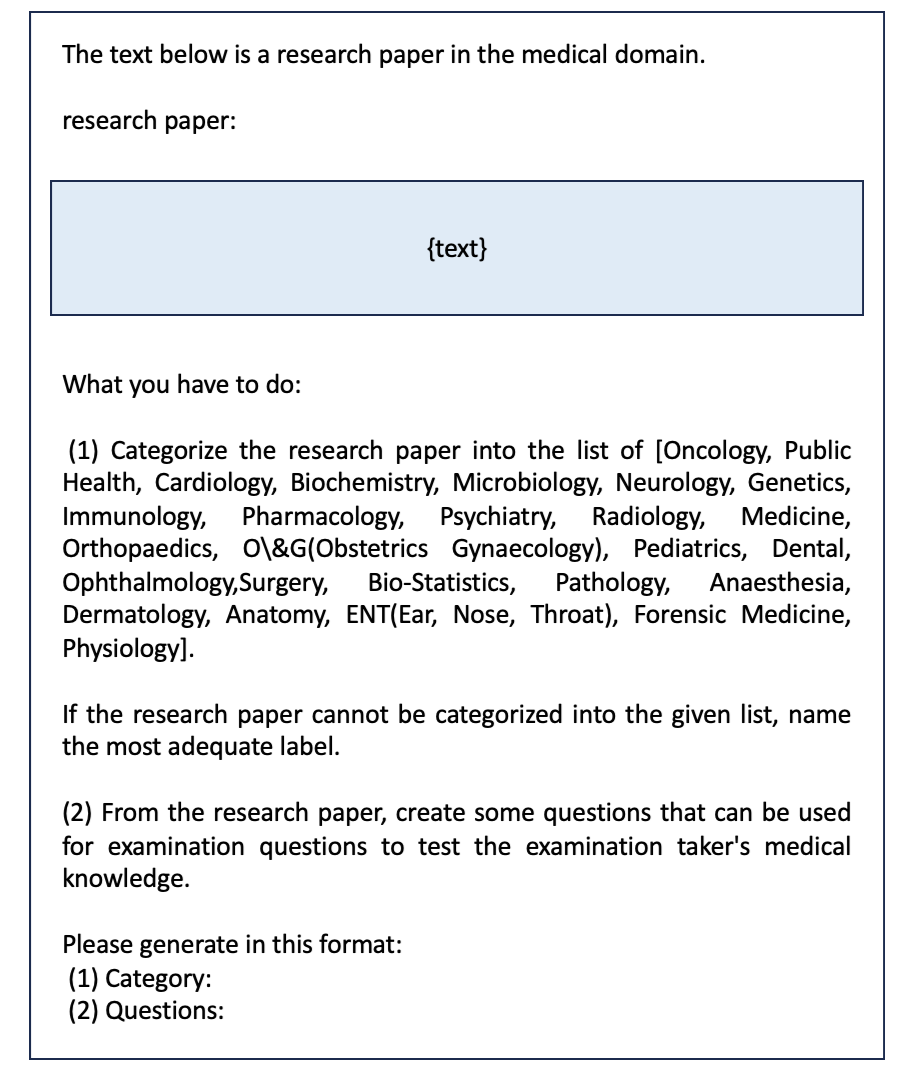} 
    \caption{Prompt used for dataset construction}
    \label{fig:prompt1}
\end{figure*}

\begin{figure*}[ht] 
    \centering
    \includegraphics[width=0.6\textwidth]{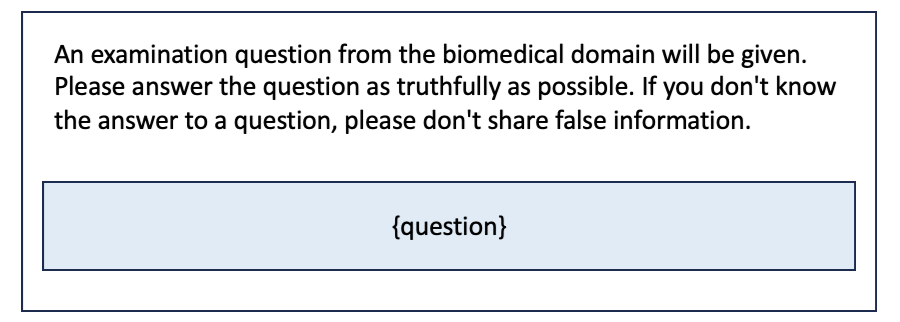} 
    \caption{Prompt used for response generation}
    \label{fig:prompt2}
\end{figure*}

\begin{figure*}[ht] 
    \centering
    \includegraphics[width=0.6\textwidth]{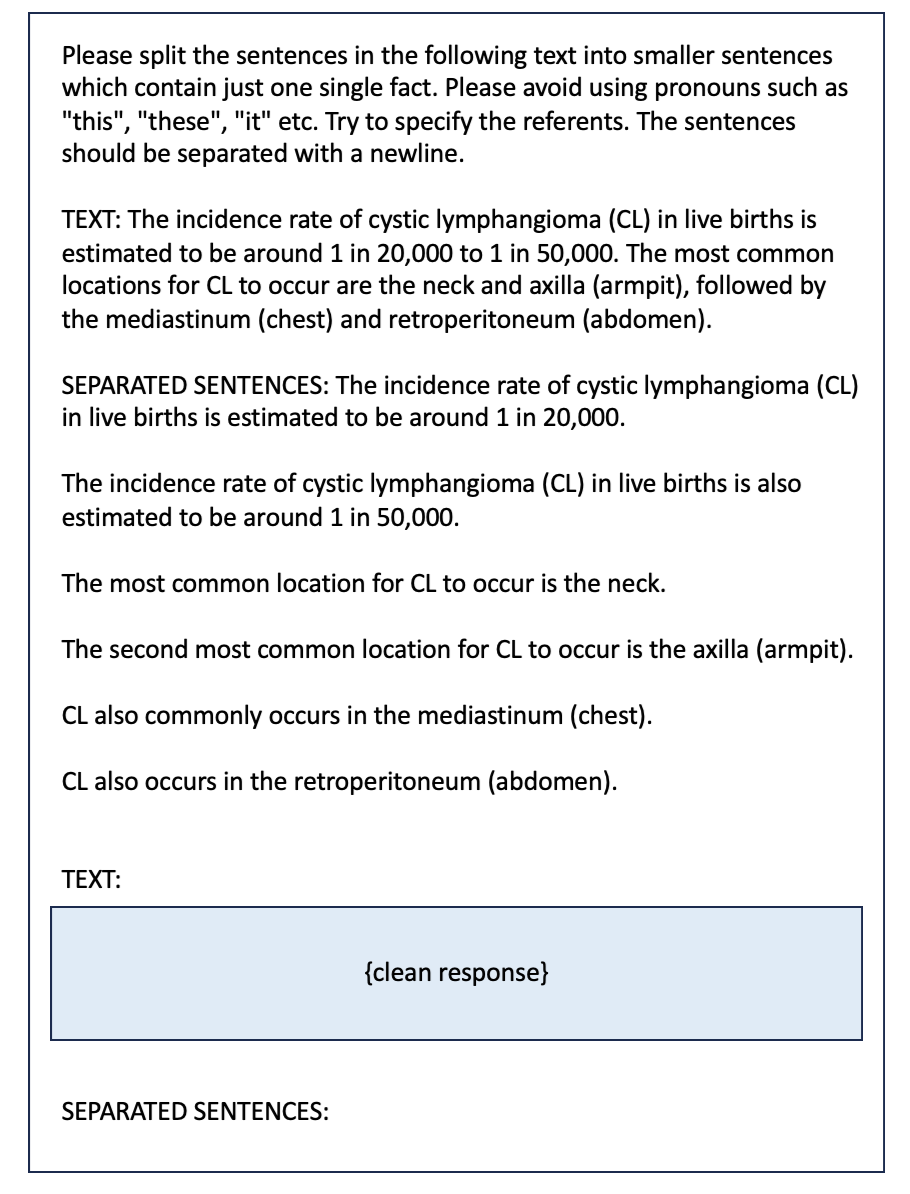} 
    \caption{Prompt used for splitting}
    \label{fig:prompt3}
\end{figure*}

\begin{figure*}[ht] 
    \centering
    \includegraphics[width=0.6\textwidth]{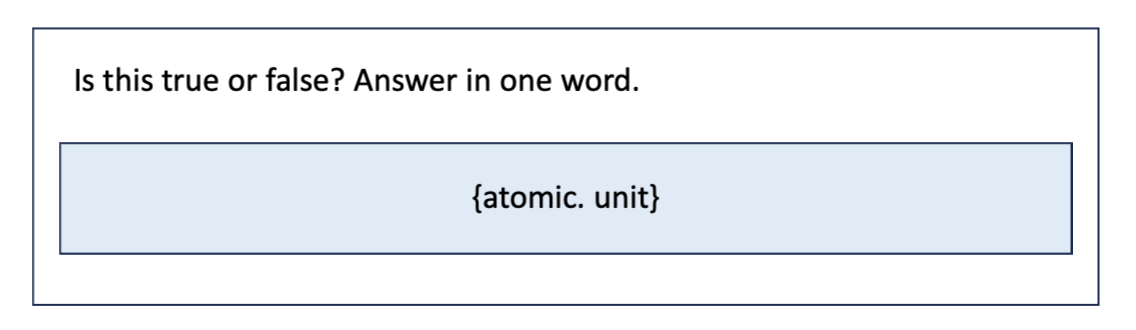} 
    \caption{Prompt used for factuality checking}
    \label{fig:prompt4}
\end{figure*}

\end{appendices}

\end{document}